\pdfoutput=1

\documentclass[11pt]{article}

\usepackage[final]{acl}

\usepackage{times}
\usepackage{latexsym}

\usepackage[T1]{fontenc}

\usepackage[utf8]{inputenc}

\usepackage{microtype}

\usepackage{inconsolata}

\usepackage{graphicx}
\usepackage{amsmath,amssymb,bm}
\usepackage {diagbox}
\usepackage[ruled, linesnumbered, titlenumbered]{algorithm2e}
\usepackage{bbm}
\usepackage{multirow}
\usepackage{booktabs}
\usepackage{pifont}

\AtBeginDocument{
  \abovedisplayskip     =0.5\abovedisplayskip
  \abovedisplayshortskip=0.5\abovedisplayshortskip
  \belowdisplayskip     =0.5\belowdisplayskip
  \belowdisplayshortskip=0.5\belowdisplayshortskip}

\title{Does Pre-trained Language Model Actually Infer Unseen Links \\in Knowledge Graph Completion?}

\author{Yusuke Sakai\textsuperscript{\dag},
Hidetaka Kamigaito\textsuperscript{\dag}, Katsuhiko Hayashi\textsuperscript{\ddag}, Taro Watanabe\textsuperscript{\dag} \\
  \textsuperscript{\dag}Nara Institute of Science and Technology \textsuperscript{\ddag}The University of Tokyo \\
  \texttt{\{sakai.yusuke.sr9, kamigaito.h, taro\}@is.naist.jp} \\
  \texttt{katsuhiko-hayashi@g.ecc.u-tokyo.ac.jp}}

\begin{document}
\maketitle
\begin{abstract}

Knowledge graphs (KGs) consist of links that describe relationships between entities. Due to the difficulty of manually enumerating all relationships between entities, automatically completing them is essential for KGs. Knowledge Graph Completion (KGC) is a task that infers unseen relationships between entities in a KG. Traditional embedding-based KGC methods (e.g. RESCAL, TransE, DistMult, ComplEx, RotatE, HAKE, HousE, etc.) infer missing links using only the knowledge from training data. In contrast, the recent Pre-trained Language Model (PLM)-based KGC utilizes knowledge obtained during pre-training, which means it can estimate missing links between entities by reusing memorized knowledge from pre-training without inference. This part is problematic because building KGC models aims to infer unseen links between entities. However, conventional evaluations in KGC do not consider inference and memorization abilities separately. Thus, a PLM-based KGC method, which achieves high performance in current KGC evaluations, may be ineffective in practical applications. To address this issue, we analyze whether PLM-based KGC methods make inferences or merely access memorized knowledge. For this purpose, we propose a method for constructing synthetic datasets specified in this analysis and conclude that PLMs acquire the inference abilities required for KGC through pre-training, even though the performance improvements mostly come from textual information of entities and relations.
\end{abstract}

\section{Introduction}

A knowledge graph (KG) is graph-structured data that includes relationships between entities as links. KGs are useful resources to inject external knowledge into NLP models. Since manually considering all possible links between entities is difficult, it is important to use a task such as KG completion (KGC), which automatically completes unseen links from seen ones in a KG.

\begin{figure}
    \centering
    \includegraphics[width=\columnwidth]{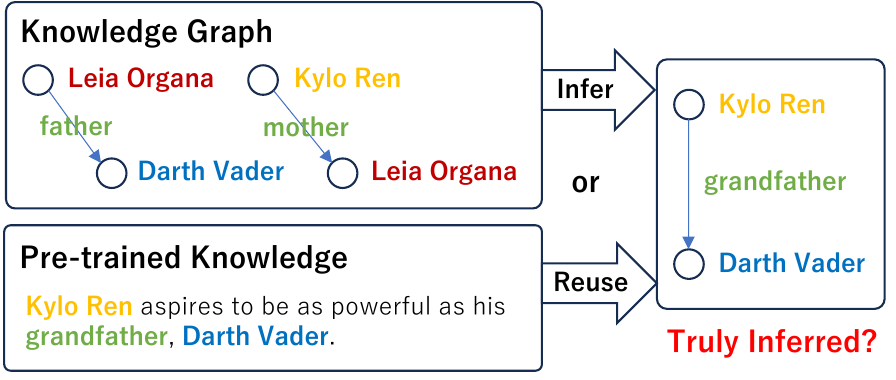}
    \caption{PLM-based KGC can reuse pre-trained knowledge of unseen links instead of inferring them.}
    \label{fig:intro}
\end{figure}

\begin{table*}[t]
\small
\resizebox{\textwidth}{!}{
\begin{tabular}{cccccccccccc}
\toprule
\multirow{5}{*}{Available Information}                             &  & \multicolumn{10}{c}{Pretrained Language Model (PLM)-based KGC}                                                                                                                                             \\
\cmidrule(lr){3-12}
& Traditional       & \multicolumn{2}{c}{\multirow{2}{*}{\textsc{Base}}}    & \multicolumn{2}{c}{\textsc{Virtual}} & \multicolumn{2}{c}{\textsc{Anonymized}} & \multicolumn{2}{c}{\textsc{Inconsistent}} & \multicolumn{2}{c}{\textsc{Fully}} \\

                                              & KGC & &     & \multicolumn{2}{c}{\textsc{World}} & \multicolumn{2}{c}{\textsc{Entities}} & \multicolumn{2}{c}{\textsc{Descriptions}} & \multicolumn{2}{c}{\textsc{Anonymized}} \\
\cmidrule(lr){3-4}\cmidrule(lr){5-6}\cmidrule(lr){7-8}\cmidrule(lr){9-10}\cmidrule(lr){11-12}
                                              & \multicolumn{1}{c}{}                                & Pre.      & Rand.       & Pre.         & Rand.          & Pre.            & Rand.             & Pre.               & Rand.                & Pre.           & Rand.           \\
                                              \midrule
Seen links in a KG                            & \textcolor{green}{\ding{51}}                                        & \textcolor{green}{\ding{51}} & \textcolor{green}{\ding{51}} & \textcolor{green}{\ding{51}}    & \textcolor{green}{\ding{51}}    & \textcolor{green}{\ding{51}}       & \textcolor{green}{\ding{51}}       & \textcolor{green}{\ding{51}}          & \textcolor{green}{\ding{51}}          & \textcolor{green}{\ding{51}}      & \textcolor{green}{\ding{51}}     \\
Descriptions of entities/relations            & \textcolor{red}{\ding{55}} & \textcolor{green}{\ding{51}} & \textcolor{green}{\ding{51}} & \textcolor{green}{\ding{51}} & \textcolor{green}{\ding{51}} & \textcolor{green}{\ding{51}} & \textcolor{green}{\ding{51}} & \textcolor{red}{\ding{55}} & \textcolor{red}{\ding{55}} & \textcolor{red}{\ding{55}} & \textcolor{red}{\ding{55}} \\
Pre-trained knowledge of KGs & \textcolor{red}{\ding{55}} & \textcolor{green}{\ding{51}} & \textcolor{red}{\ding{55}} & \textcolor{red}{\ding{55}} & \textcolor{red}{\ding{55}} & \textcolor{red}{\ding{55}} & \textcolor{red}{\ding{55}} & \textcolor{red}{\ding{55}} & \textcolor{red}{\ding{55}} & \textcolor{red}{\ding{55}} & \textcolor{red}{\ding{55}} \\
Abilities obtained by pre-training & \textcolor{red}{\ding{55}} & \textcolor{green}{\ding{51}} & \textcolor{red}{\ding{55}} & \textcolor{green}{\ding{51}} & \textcolor{red}{\ding{55}} & \textcolor{green}{\ding{51}} & \textcolor{red}{\ding{55}} & \textcolor{green}{\ding{51}} & \textcolor{red}{\ding{55}} & \textcolor{green}{\ding{51}} & \textcolor{red}{\ding{55}} \\
\bottomrule
\end{tabular}}
\caption{Available information for each configuration. When compared, we can reveal what improves the KGC performance on PLMs. \textsc{Base} denotes the setting on the original data, and \textsc{Virtual World} (\S\ref{sec:shuffle}), \textsc{Anonymized Entities} (\S\ref{sec:random}), \textsc{Inconsistent descriptions} (\S\ref{sec:shuffle-descriptions}), and \textsc{Fully Anonymized} (\S\ref{sec:random-descriptions}) denote the settings on our synthetic datasets. Pre. and Rand. denote the setting with pre-trained and randomly initialized weights, respectively.}
\label{tab:configurations}
\end{table*}

As a basic method for KGC, KG embedding (KGE) is a popular chioce for this task. KGE embeds entities and their relationships as continuous vectors and then calculates the plausibility of unseen links.
Traditional KGE methods learn these embeddings only from a target KG \cite{rescal,transE,distmult,complex,rotate,hake,house}. Thus, they purely infer unseen links to complete KGs.

Similar to other NLP fields, KGC also utilizes pre-trained language models (PLMs) \citep{kg-bert, pkgc, shen-etal-2022-lass, knnkge, mem-kgc, openworld-kgc, star, wang-etal-2021-kepler, wang-etal-2022-simkgc, genkgc, kgt5, chen-etal-2022-s2skgc, xie-etal-2023-lambdakg, zhu2023autokg}. Unlike traditional KGE methods, PLM-based KGE methods can access knowledge obtained through pre-training. This characteristic makes PLM-based KGE methods achieve higher KGC performance than the traditional KGE methods.

However, since the purpose of KGC is to infer unseen links from seen links in KGs, we should separately consider the performance gain from reusing the information of the unseen links obtained in pre-training and inferring unseen links from the seen links in KGs. Figure \ref{fig:intro} shows an example of PLM-based KGC. As we can see, PLM-based KGC methods can estimate unseen links without inferring them from seen links in the target KG. This characteristic is problematic because we cannot estimate the inference ability of PLM-based KGC methods for truly unseen relationships between entities in KGs.

To address this issue, we propose a method to create synthetic datasets for KGC tasks intended to separately evaluate KGC performance by reusing the knowledge from pre-training corresponding to target unseen links and inferring from seen links in KGs. More specifically, we change the textual information of entities and relations while maintaining the graph structure of KGs, thereby creating an environment different from the PLMs’ knowledge corresponding to unseen links in KGs. Due to this change, PLMs cannot rely on their pre-trained knowledge and must rely on their pure inference abilities. Table \ref{tab:configurations} summarizes the configurations provided by our synthetic datasets. By comparing these configurations, we can reveal what actually contributes to the KGC performance of PLMs.

We conducted experiments on various pre-trained models under our controlled synthetic dataset constructed from WN18RR~\cite{WN18RR}, FB15k-237~\cite{FB15k-237}, and Wikidata5m~\cite{wang-etal-2021-kepler}. The results showed that PLMs acquire the inference abilities required for KGC in pre-training but rely more on textual information of entities and relations in KGs. We also observed that the KGC performance of PLM-based KGC without pre-trained information is comparable to or lower than that of TransE, the traditional KGC. This finding indicates the importance of both traditional and PLM-based KGC methods.

\section{Knowledge Graph Completion}

\begin{figure*}[t]
\centering
\includegraphics[width=\textwidth]{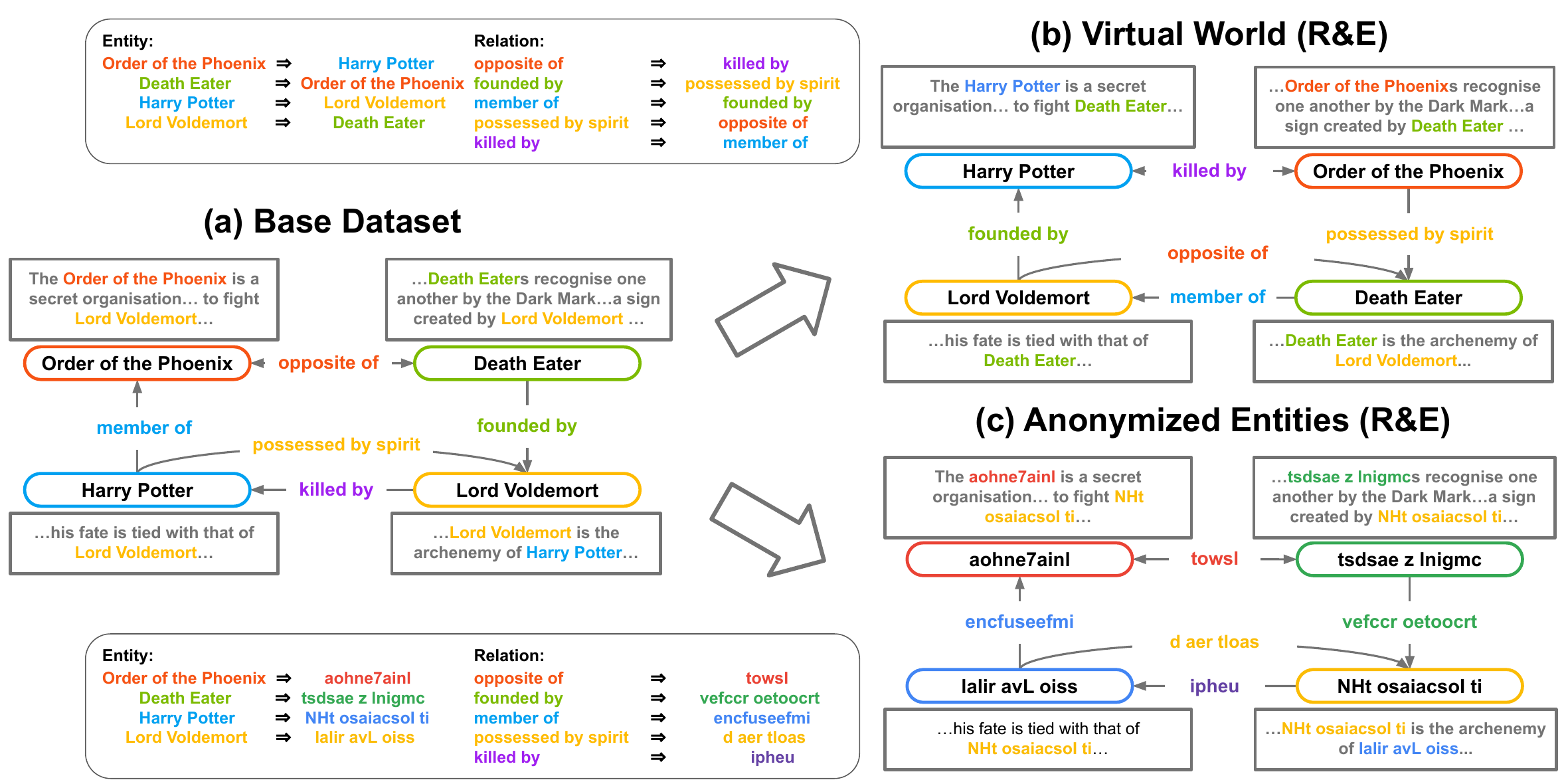}
\caption{(a): Example of a KG with entity descriptions for PLM-based methods. Each entity has a corresponding description.
(b) and (c) are the datasets used in this study. We primarily apply two methods for creating these datasets in \textsc{Virtual World} (\S\ref{sec:shuffle}) and \textsc{Anonimized Entities} (\S\ref{sec:random}). (b) described in \textsc{Virtual World} (\S\ref{sec:shuffle}) involves swapping the names assigned to entities and relations in the base dataset respectively. (c) described in \textsc{Anonimized Entities} (\S\ref{sec:random}) substitutes the names of entities and relations in the base dataset with random strings. Note that in both procedures, any entities appearing within the description text are replaced with their corresponding transformed names to maintain the graph structure within the descriptions.}
\label{fig:main}
\end{figure*}

\subsection{Task Definition for KGs with Descriptions}
\label{subsec:task_definition}

We assume that a KG $\mathcal{G}$ includes descriptions defined as a tuple, $\mathcal{G}=(\mathcal{E}, \mathcal{R}, \mathcal{T}, \mathcal{D})$, where $\mathcal{E}$ denotes a set of entities, $\mathcal{R}$ denotes a set of relations, $\mathcal{T}$ denotes a set of triples, and $\mathcal{D}$ denotes descriptions for the entities.
Each triple is represented as $(h, r, t) \in \mathcal{T}$, where $h$ and $t \in \mathcal{E}$ are the head and tail entities, respectively, and $r \in \mathcal{R}$ is the relation. 
Every entity $e_i \in \mathcal{E}$ has a corresponding description $d_i \in \mathcal{D}$.
KGC is a task to fill in the missing triples in KGs. Specifically, this involves using a query, a partial triple $(h, r, ?)$ or $(?, r, t)$ to predict its answer, an entity at the position of $?$, within the KG. 
Note that the prediction is exclusively focused on entities; predicting their corresponding descriptions is not required.

KGC is often evaluated by rank prediction metrics such as Hits@$k$~($k \in \{1,3,10\}$), mean rank (MR), and mean reciprocal rank (MRR). Hits@$k$ calculates the proportion of correct entities ranked among the top-$k$, MR is the average rank of all test triples, and MRR is the average reciprocal rank of all test triples. 

\subsection{KGC Methods}
\label{subsec:kgc_methods}

Traditional KGC methods, e.g., RESCAL \citep{rescal}, TransE \citep{transE}, DistMult \citep{distmult}, ComplEx \citep{complex}, RotatE \citep{rotate}, HAKE \citep{hake}, and HousE \citep{house}, primarily focus on the structure of KGs, without considering the extensive textual information.

However, recent advancements integrating PLMs have allowed KGC methods to encode text~\cite{kg-bert, pkgc, shen-etal-2022-lass, knnkge, mem-kgc, openworld-kgc, star, wang-etal-2021-kepler, wang-etal-2022-simkgc} or generate facts~\cite{genkgc, kgt5, chen-etal-2022-s2skgc, xie-etal-2023-lambdakg, zhu2023autokg}, thereby enhancing the KGC performance. These methods can be broadly divided into two categories based on their usage: discrimination-based methods that utilize PLM encoders, and generation-based methods that utilize PLM decoders~\cite{pan2023unifying} (see Appendix \ref{sec:plm-based-methods} for the details).

\section{Synthetic Dataset Construction}
\label{sec:how-to-create}

To analyze the behavior of PLM-based KGC methods, we create synthetic data corresponding to each setting in Table \ref{tab:configurations}. These settings affect the usable information of the PLM-based KGC methods but do not influence the traditional KGE methods. We explain the details for each setting in the following subsections.

\begin{algorithm}[t]
\small
\SetAlgoLined
\KwData{Input array $arr$ of size $n$, Set of removed edges $removed\_edges$}
\KwResult{Generated array $res$}
Create an empty graph $G$\;
\For{$i \leftarrow 0$ \KwTo $n-1$}{
  \For{$j \leftarrow 0$ \KwTo $n-1$}{
    \If{$arr[i] \neq arr[j]$ and $(arr[i], arr[j])$ is not in $removed\_edges$}{
      add edge $(i, n + j)$ in $G$\;
    }
  }
}
$match$ $\leftarrow$ maximum matching($G$) \\
$res \leftarrow$ an empty list of size $n$\;
\For{$i\leftarrow 0$ \KwTo $n-1$}{
  $index \leftarrow match[i] - n$\;
  $res[i] \leftarrow arr[index]$\;
}
\Return{$res$}
\caption{Derangement by Bipartite Graph}
\label{algorithm:bipartite}
\end{algorithm}

\subsection{Virtual World}
\label{sec:shuffle}

To separate the pre-trained knowledge of PLMs and a target KG, we create a virtual world by shuffling each entity and/or relation name in the KG.

As shown in Figure \ref{fig:main}(b), we shuffle the textual information associated with each entity and/or relation while keeping the graph structure within the created synthetic dataset. 
To ensure there are no un-shuffled elements, we shuffle the entities using the derangement algorithm by \citet{derangement}.

However, there are dramatically fewer relations compared to entities (e.g., ten relations for ten thousand entities), and if relations are shuffled, the triple remains unchanged in many cases.\footnote{In the case of (Johann Bernoulli, wasBornIn, Basel) and (Johann Bernoulli, diedIn, Basel), the swapping of the relations wasBornIn and diedIn does not change the triples.}
To address these cases, we apply a derangement based on a bipartite graph~\cite{bipartite2019, bipartite2020} in Algorithm~\ref{algorithm:bipartite} for relations.

In Algorithm~\ref{algorithm:bipartite}, we introduce  $removed\_edges$, a set to the bipartite graph-based derangement. 
Lines 4--6 in Algorithm~\ref{algorithm:bipartite} delete edges leading to multiple relations in a triplet $(h,*, t)$, thereby preventing transitions to these relations.\footnote{If $removed\_edges$ is empty, it is a normal derangement.}
We use the Hopcroft-Karp algorithm~\citep{Hopcroft-Karp} for maximum bipartite matching.

Additionally, we use Trie search~\cite{marisa-trie} to comprehensively search for entity representations within each description in Figure~\ref{fig:main} and change them into their post-shuffled text representations. This procedure treats the relationships between entities within the descriptions while maintaining their original graph structure in the descriptions.

\subsection{Anonymized Entities}
\label{sec:random}

\textsc{Virtual World} can separate the pre-trained knowledge of PLMs and a target KG. However, this setting may underestimate the KGC performance caused by the overwrap of the entity and/or relation names between pre-trained knowledge and the target KG.  

The \textsc{Anonymized Entities} setting can solve this problem by replacing the textual information associated with each entity and/or relation with a random string while keeping the original graph structure within the dataset, as in Figure\ref{fig:main}(c). Afterward, we also replace the entity representations within the description with these random strings using Trie search, the same as \textsc{Virtual World}~(\S\ref{sec:shuffle}).

Since the random strings should follow language characteristics, we first construct character-level unigram language models $P(s_i)$, including space characters from the set of textual information of each entity and relation.

Next, we generate random strings $\boldsymbol{s} = s_1, s_2,$ $\ldots, s_n$ based on the character-level unigram language model $p(\boldsymbol{s})$, i.e., the product of the probabilities of unigram character in the strings:
\begin{equation}
    p(\boldsymbol{s})= \prod^n_{i=1}p(s_i).
\end{equation}
We stop the generation of strings when an end-of-sequence symbol is sampled. The strings are treated as a series of independent characters, allowing us to generate entirely random strings without using information about co-occurrence between characters. 
However, we preserve information for the randomly sampled sequences across the entire dataset so that each entity or relation is replaced with a unique sequence avoiding duplicates.

\subsection{Inconsistent Descriptions}
\label{sec:shuffle-descriptions}

\begin{figure}
\centering
\includegraphics[width=0.48\textwidth]{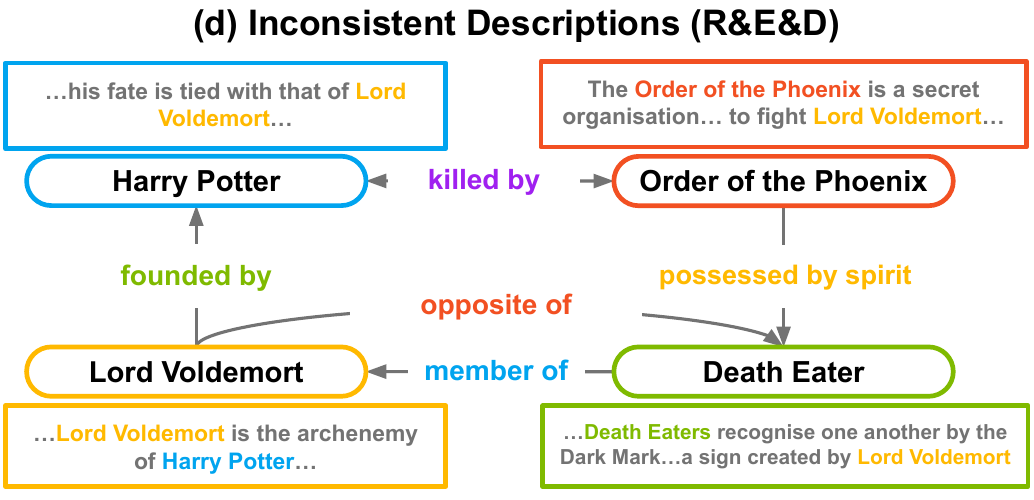}
\caption{Example of a synthetic dataset created in \textsc{Inconsistent Descriptions}~(\S\ref{sec:shuffle-descriptions}). Compared to Figure\ref{fig:main}(b) which shows an example of \textsc{Virtual World}~(\S\ref{sec:shuffle}), the descriptions here also move to the same positions as the entities. Also, the entities in the descriptions do not change. At first glance, it appears the description explains the real-world relationships of the corresponding entities, but the relationships between entities within the synthetic dataset are actually broken.}
\label{fig:shuffle-descriptions}
\end{figure}

To measure the effect of descriptions on PLM-based KGC, we isolate the entity and relation knowledge from the description by breaking the consistency between the graph structure and descriptions in addition to the shuffle of entity and/or relation names.

\textsc{Inconsistent Descriptions} has two variations, one in which only the descriptions are shuffled and the other in which both the descriptions and entities/relations are shuffled.
In the first variation, we derive the scenario in which there is no correspondence between an entity and its description by shuffling the set of descriptions via a derangement to get a new set $d' \in D'$. 
Then, we assign for each entity the new descriptions from $D'$, i.e., $\forall e_i \in E, e_i : d_i \rightarrow d'_i$.

The second variation considers the descriptions and entities presented in  Figure~\ref{fig:shuffle-descriptions}.
The difference from  Figure~\ref{fig:main}(b) for \textsc{Virtual World}~(\S\ref{sec:shuffle}) lies in the way it handles the descriptions.
In \textsc{Inconsistent Descriptions}, descriptions are also shuffled together with the corresponding textual information when performing \textsc{Virtual World}, but the entities in the descriptions are preserved.
In other words, when we map from $e_i$ to $e_j$, we similarly map from $d_i$ to $d_j$. 

Even though the descriptions explain the entities in the real world, they diverge from the relationships among entities in the dataset after the shuffle operation. Thus, if the model relies too much on the descriptions, it will be confused by this inconsistency.

\subsection{Fully Anonymized}
\label{sec:random-descriptions}

\begin{figure}
\centering
\includegraphics[width=0.48\textwidth]{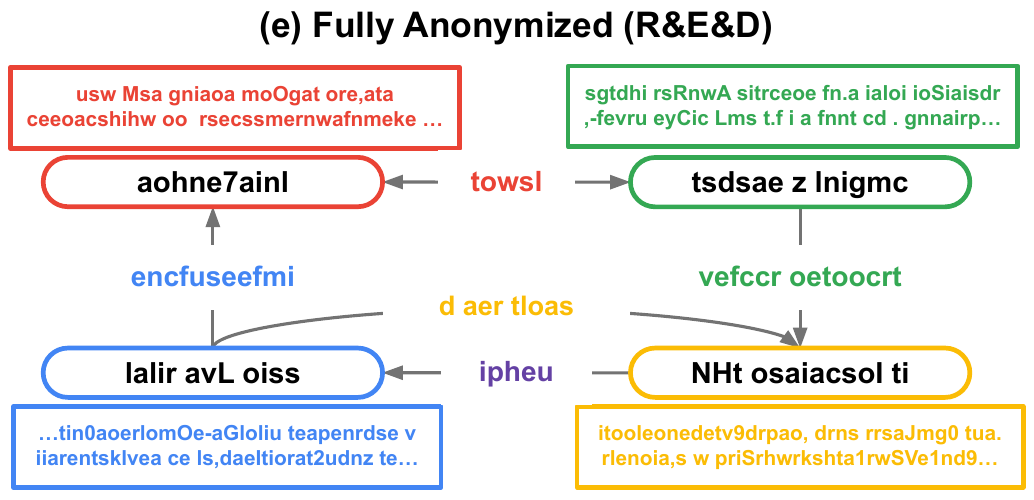}
\caption{Example of a synthetic dataset created in \textsc{Fully Anonimized}~(\S\ref{sec:random-descriptions}). Compared to Figure \ref{fig:main}(c), which shows an example of \textsc{Anonymized Entities}~(\S\ref{sec:random}), the descriptions are here also changed into random strings. The descriptions become noisy information, and it becomes impossible to utilize any information from them.}
\label{fig:random-descriptions}
\end{figure}

Figure~\ref{fig:random-descriptions} shows an example of \textsc{Fully Anonymized}, which is similar to \textsc{Anonymized Entities} (\S\ref{sec:random}) in Figure~\ref{fig:main}(c) but differs in whether or not there is an operation on the descriptions.
We replace the descriptions with random strings
using the character-level unigram model utilized in \textsc{Anonymized Entities}~(\S\ref{sec:random}), while we keep the original structure of the KGs.
This setting aims to mitigate underestimating the KGC performance caused by the overlap of the entity and/or relation names between pre-trained knowledge and the target KG.
Note that the random string generation is applied independently to entities, relations, and descriptions.
The key difference between \textsc{Fully Anonymized} and \textsc{Inconsistent Descriptions} (\S\ref{sec:shuffle-descriptions}) lies in whether the descriptions are readable sentences or not; if they are not, the PLMs in \textsc{Fully Anonymized} cannot rely on any pre-trained knowledge.

\section{Experiments}
\subsection{Settings}

\paragraph{Metrics} We analyze how the inference capabilities are affected by each synthetic dataset (\S\ref{sec:how-to-create}) measured with the Hits@10 metric on the test dataset and the validation dataset in the KGC task.\footnote{We also measured Hits@1, Hits@3, MRR, and MR, and all showed similar trends. In this paper, we present the results using hits@10 for brevity.}

\begin{table}
\centering
\footnotesize
\setlength{\tabcolsep}{1.2pt}
\begin{tabular}{l|ccccc}
\hline
Dataset & \#entity & \#relation & \#train & \#valid & \#test\\
\hline
WN18RR & 40,943 & 11 & 86,835 & 3,034 & 3,134 \\
FB15k-237 & 14,541 & 237 & 272,115 & 17,535 & 20,466 \\
Wikidata5m & 4,594,485 & 822 & 20,614,279 & 5,163 & 6,894 \\

\hline
\end{tabular}
\caption{Dataset statistics.}
\label{tab:datasets}
\end{table}

\begin{figure*}[t]
\centering
\includegraphics[width=\textwidth]{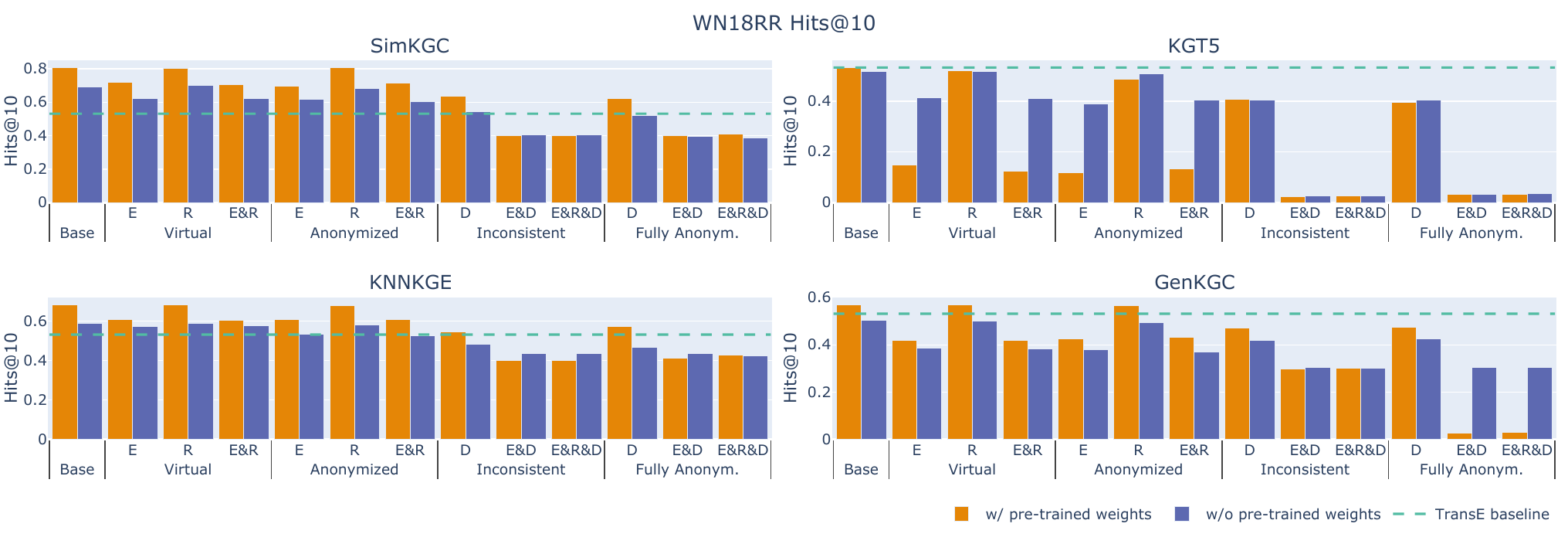}
\caption{The hits@10 results on WN18RR. ``E'', ``R'', and ``D'' represent entity, relation, and description, respectively. For example, ``E\&R'' denotes the application of the method to both entities and relations. For comparison, we have also included the hits@10 results on WN18RR by TransE reported by \citet{nathani-etal-2019-learning}, which are the same score because the TransE model does not require textual information. The graphs on the left represent Discrimination-Based Methods, while those on the right represent Generation-Based Methods.}
\label{fig:WN18RR-hits10}
\end{figure*}

\begin{figure*}[t]
\centering
\includegraphics[width=\textwidth]{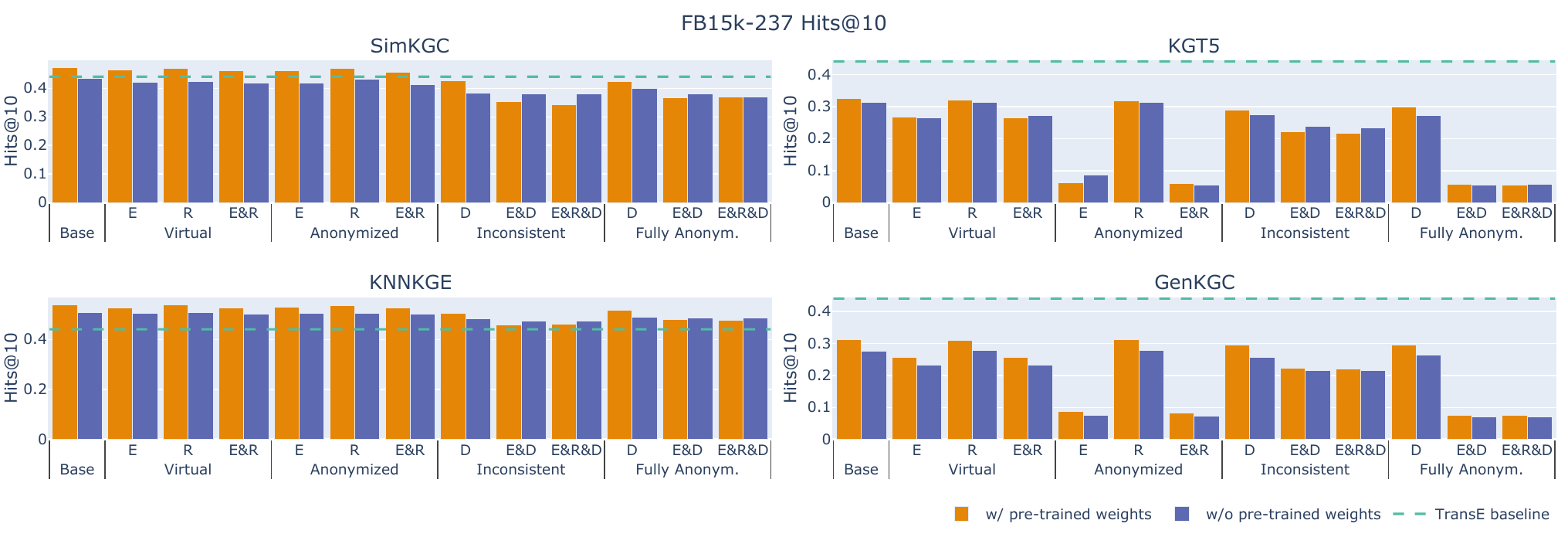}
\caption{Hits@10 results on FB-15k-237. The supplementary explanation is the same as in Figure \ref{fig:WN18RR-hits10}.}
\label{fig:FB15k-237-hits10}
\end{figure*}

\paragraph{Datasets}
We used WN18RR, FB15k-237, and  Wikidata5m\footnote{We follow the transductive setting in \citet{wang-etal-2021-kepler}.} as the base datasets; the details are shown in Table \ref{tab:datasets}.\footnote{We use the datasets with textual information provided by \citet{kg-bert} for WN18RR, FB15k-237, and by \citet{wang-etal-2021-kepler} for Wikidata5m.} We applied \textsc{Virtual World} (\S\ref{sec:shuffle}) and \textsc{Anonymized Entities} (\S\ref{sec:random}) to the entities and/or relations for creating synthetic datasets, resulting in a total of six types of datasets.
Furthermore, we applied \textsc{Inconsistent Descriptions} (\S\ref{sec:shuffle-descriptions}) with and without \textsc{Virtual World} for entities and/or relations. \textsc{Inconsistent Descriptions} (\S\ref{sec:random-descriptions}) is also applied with and without \textsc{Anonymized Entities}, and thus, we obtained additional six types of datasets.
In total, we have 13 types of datasets, including the original one for each base dataset.

\paragraph{Comparison Methods} We employ SimKGC \cite{wang-etal-2022-simkgc} and kNN-KGE \cite{knnkge} as Discriminative-based methods, and KGT5~\cite{kgt5} and GenKGC~\cite{genkgc} as Generation-based methods. We use the LambdaKG framework~\cite{xie-etal-2023-lambdakg} as the base implementation, with hyper-parameters set to their default values. The seed value is fixed for all experiments.\footnote{We conducted pilot studies with various seeds for several datasets and models. The variance observed was around 0.02, so a fixed seed value was chosen. For example, the Hits@10 scores in kNN-KGE on WN18RR applied with \textsc{Fully Anonymized} (\S\ref{sec:random-descriptions}) to all descriptions, entities, and relations were 0.426 ± 0.001 with three different seeds.} We set early stopping for WN18RR and FB15k-237 when the Hits@10 value on the validation data did not improve for four epochs. For Wikidata5m, we conducted training only one epoch.\footnote{We only report the results from SimKGC, as kNN-KGE could not be executed due to computational resource limitations, and both KGT5 and GenKGC did not produce scores under these settings. We conducted all experiments on a single NVIDIA A100 (40GB) or a single NVIDIA A6000 (48GB).}
We also compare two cases: using pre-trained weights and setting weights randomly.

\subsection{Results and Analysis}

\begin{figure}[t]
\centering
\includegraphics[width=0.48\textwidth]{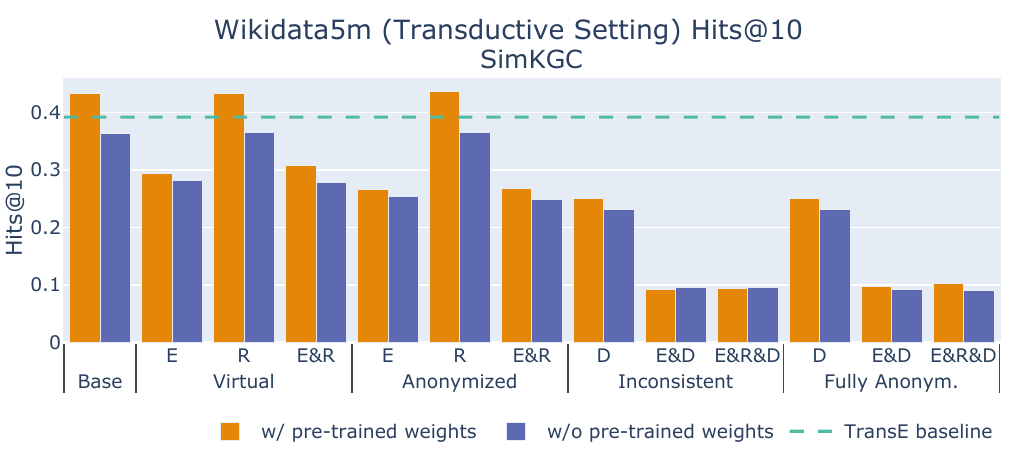}
\caption{Hits@10 results on Wikidata5m by SimKGC. We have also included the Hits@10 results on WN18RR by TransE reported by \citet{wang-etal-2021-kepler}.
}
\label{fig:wikidata5m-hits10}
\end{figure}

\begin{figure}[t]
\centering
\includegraphics[width=0.48\textwidth]{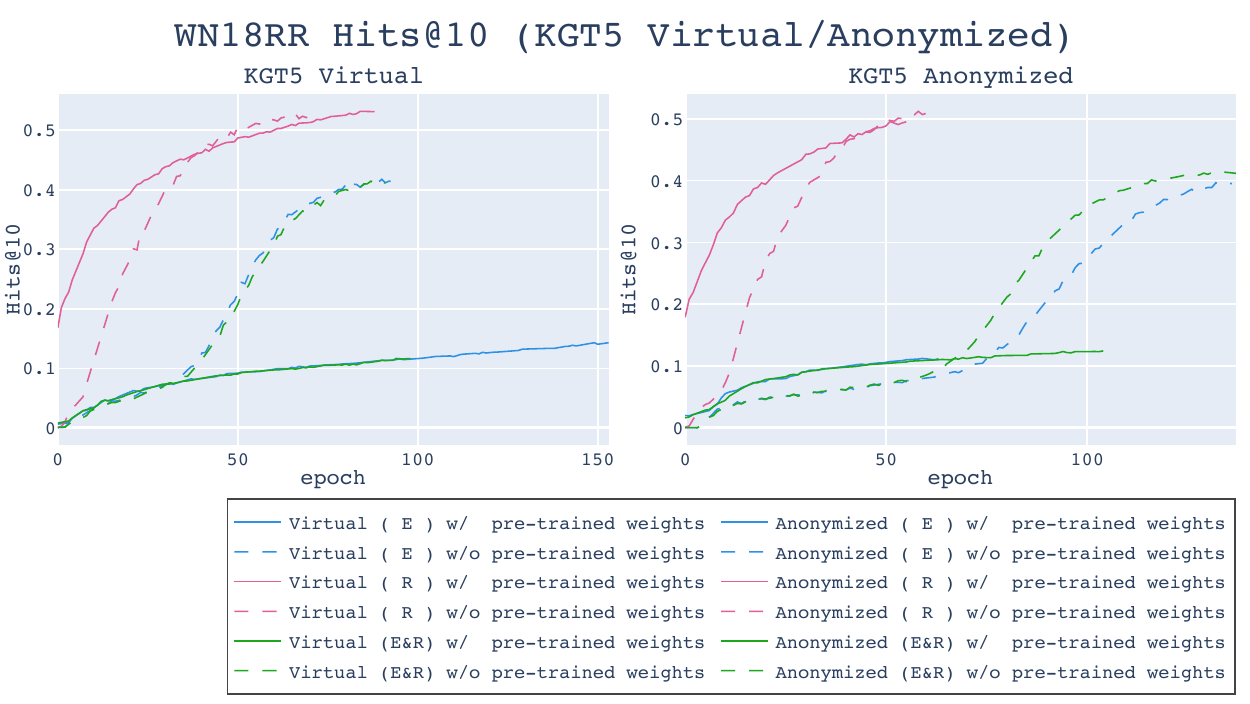}
\caption{The plots show Hits@10 scores on WN18RR for the validation data at each epoch. The solid line represents using pre-trained weights, and the dashed line represents initializing weights randomly.}
\label{fig:WN18RR-eval-short}
\end{figure}

\begin{figure*}[t]
\centering
\includegraphics[width=0.99\textwidth]{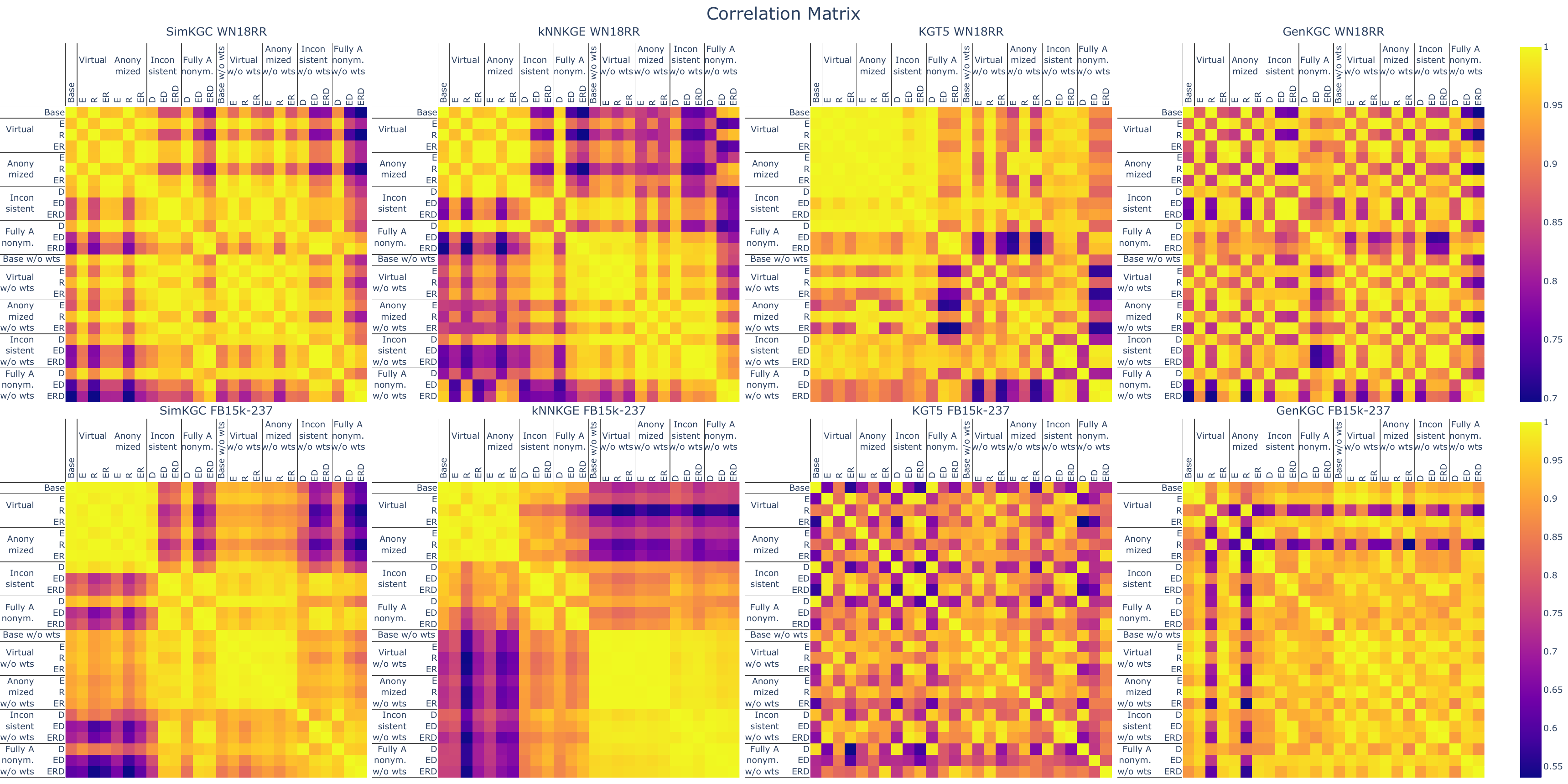}
\caption{The correlation matrix (Pearson's correlation) shows the hits@10 values for the validation data for each dataset and each model.``Virtual'', ``Anonymized'', ``Inconsistent'', and ``Fully Anonym.'' represent the methods applied in Sections \ref{sec:shuffle}, \ref{sec:random}, \ref{sec:shuffle-descriptions}, and \ref{sec:random-descriptions}, respectively. ``E'', ``R'', and ``D'' represent entity, relation, and description, respectively. For example, ``ER'' denotes the application of the method to both entities and relations. ``w/o wts'' means training from scratch with random initial values. The two graphs on the left are Discrimination-Based Methods, and the two on the right are Generation-based Methods.}
\label{fig:correlation}
\end{figure*}

\begin{table*}[t]
    \centering
    \small
    \setlength{\tabcolsep}{5.5pt}
    \begin{tabular}{c|c|c|c|c|c|c|c|c|c|c|c|c}
    \hline
   \multirow{2}{*}{\#Relation}   &\multicolumn{4}{c|}{WN18RR (\%)} & \multicolumn{4}{c|}{FB15k-237 (\%)} & \multicolumn{4}{c}{Wikidata5m (\%)} \\ \cline{2-13}
      & Train & Valid & Test & Total & Train & Valid & Test & Total & Train & Valid & Test & Total \\ \hline
    1 & 60.09 & 97.55 & 97.51& 57.68 & 13.52 & 65.26 & 61.64& 12.56 & 24.23 & 98.58 & 98.73& 24.22 \\
    2 & 35.39 &  2.41 & 2.49 & 37.31 & 14.02 & 24.44 & 25.61& 12.98 & 17.23 &  1.98 &  0.93& 17.21 \\
    3 &  4.21 &  0.02 &   --  &  4.61 & 11.39 &  7.39 &  8.90& 10.88 & 21.88 &  0.21 &  0.28& 21.88 \\
    4 &  0.30 &  0.02 &   --  &  0.38 & 10.01 &  2.08 &  2.67&  9.53 & 11.95 &  0.01 &  0.05& 11.95 \\
    5 &  0.02 &   --   &   --  &  0.02 &  9.06 &  0.51 &  0.76&  8.85 &  7.94 &  0.01 &  0.01&  7.94 \\
 Over &  --    &   --   &   --  &   --   & 42.00 &  0.31 &  0.41& 45.19 & 16.78 &   --   &   --  & 16.79 \\ \hline
    \end{tabular}
    \caption{The number of relations assigned to each entity in each dataset. Note that some entities may be associated with multiple entities under certain entity and relation queries.}
    \label{tab:dataset_stats}
\end{table*}

\subsubsection{Effect of knowledge in PLMs}
\label{sec:The-Effects-of-PLMs-Knowledge}
The results for each model and dataset on WN18RR and FB15k-237 are shown in Figures~\ref{fig:WN18RR-hits10} and \ref{fig:FB15k-237-hits10}, and the results from SimKGC on Wikidata5m are shown in Figure~\ref{fig:wikidata5m-hits10}. In the ``Base'' setting, all models with the pre-trained weights were better than those without them. When the models are trained without pre-training weights, they have to infer unseen links based only on information within the training data of the KGC dataset.

Comparing ``Base'', ``Virtual'', and ``Anonymized'' settings, we can see performance degradations by restricted access to knowledge for entity names obtained in pre-training. However, the models without the pre-trained weights achieved better or at least comparable results, especially when changes were made to both entities and their descriptions, as you can see in the ``Inconsistent'' and ``Fully Anonym.'' settings. From the result, we hypothesize that the performance gain by pre-trained weights in ``Virtual'' and ``Anonymized'' settings comes from the pre-trained ability to read textual information.

Figure~\ref{fig:wikidata5m-hits10} shows the importance of pre-trained knowledge for entity names in Wikidata5m. For the further analysis, we applied the interquartile range (IQR), an outlier detection method~\cite{IQR-outlier}, and the result show the significant performance gap between models with and without pre-trained weights only when entity names and their descriptions were unchanged. This finding indicates that PLM knowledge significantly contributes to the model's inference, especially in Wikidata5m.

\subsubsection{Biases caused by PLM knowledge on inference for unseen links}

We discussed the benefit of PLM knowledge in Section \ref{sec:The-Effects-of-PLMs-Knowledge}, but on the other hand, PLM knowledge may adversely affect the inference for unseen entities. Especially in Figures~\ref{fig:WN18RR-hits10} and \ref{fig:FB15k-237-hits10}, it is clear that the difference between with and without pre-trained knowledge significantly affected the scores, particularly in the case of entity changes in KGT5. 

Figure~\ref{fig:WN18RR-eval-short} shows the training curves of Hits@10 on WN18RR for the validation data. Remarkable results were observed for the \textsc{Virtual World} and \textsc{Anonymized Entities} methods in KGT5: namely the models using pre-trained weights could not learn well, even with sufficient epochs of training, whereas the models without pre-trained weights exhibited inference capability for unknown entities. These results suggest that while PLM knowledge helps infer unseen links, it may prevent the learning of new relationships due to the relationships included in the PLM knowledge.

\subsubsection{Which factors (entity, relation, description) affect inference ability?}
\label{sec:which-factors}

Figure~\ref{fig:correlation} shows the correlation matrix of Hits@10 scores on the validation data for each dataset and model. 
In Figures~\ref{fig:WN18RR-hits10},~\ref{fig:FB15k-237-hits10},~and~\ref{fig:correlation}, the results from the base dataset and changes to relations indicate strong correlations in the learning process and Hits@10 scores in the test data. Therefore, the model is not affected by changes to relations when inferring unseen links. 
As shown in Table \ref{tab:datasets}, the number of relations is significantly smaller than that of entities. Moreover, Table \ref{tab:dataset_stats} reveals entities with only one assigned relation in the KGC dataset: 12\% in FB15k-237 and over 50\% in WN18RR. 
This suggests that the models can infer connections between entities without considering their actual relations. 

Figure \ref{fig:correlation} also shows a correlation between \textsc{Virtual World} and \textsc{Anonymized Entities}, indicating that which kind of textual information is used for inference is less important than than the consistency in relationships between entities in each triplet. 
Additionally, when changing both the entity and the description, the score decreases in Figures \ref{fig:WN18RR-hits10} and \ref{fig:FB15k-237-hits10}. 
Table \ref{tab:entity_in_descriptions} shows how many entities to predict are included in the description of query entities; in WN18RR, about 15 \% of the entities may be able to solve the KGC task just by extracting information from the description. Changes to the description only are less likely to be affected, but changing both the entity and the description eliminates clues to the answer from both, leading to a decrease in the inference capabilities with PLM.

\subsubsection{Effect of model structures on performance}

When comparing Generation-based methods with Discrimination-based methods, the former are substantially affected by random strings of entities. As shown in Figure \ref{fig:WN18RR-hits10}, KGT5 and GenKGC without the pre-trained weights learn better than those that have them.
Furthermore, Figure~\ref{fig:WN18RR-eval-short} shows that scores do not improve even with sufficient training, which suggests that the difference in scores is not due to the early stopping. Thus, PLM knowledge prevents learning new relationships from descriptions in Generation-based methods.

\citet{kwon-etal-2023-hierarchical} point out a benefit of predicting structured labels by Generation-based methods is handling the relationship of labels through implicitly infused label embeddings \citep{xiong-etal-2021-fusing,zhang-etal-2021-language} on the decoder. However, the current usage of Generation-based methods in KGC only predicts a single entity without its description for each query. Therefore, in the current usage, Generation-based methods cannot handle relationships between entities and consider their description information.

\begin{table}[t]
    \centering
    \small
    \setlength{\tabcolsep}{3pt}
    \resizebox{\columnwidth}{!}{
    \begin{tabular}{c|c|c|c|c} \hline
    & Train (\%) & Valid (\%) & Test (\%) & Total (\%) \\ \hline
    WN18RR & 15.03 & 15.62 & 15.34 & 15.06 \\
    FB15k-237 & 6.11 & 4.68 & 4.50 & 5.92 \\
    Wikidata5m & 4.58 & 4.99 & 4.58 & 4.58\\ \hline
    \end{tabular}}
    \caption{Percentage of target entities to predict is included in the description of the query entity for each dataset. These triplets can be solved by simply extracting information from the descriptions without performing any inference in the KGC tasks.}
    \label{tab:entity_in_descriptions}
\end{table}

Moreover, Generation-based methods are influenced by the string of the output entity, as seen in Figures from \ref{fig:WN18RR-hits10} to \ref{fig:wikidata5m-hits10}. 
On the other hand, Discrimination-based methods are less affected by the textual information, in contrast to Generation-based methods that are affected by random strings that lack the characteristics of language and are thus unsuitable for generation (see Appendix \ref{sec:inference-capability} for further analysis).

\section{Related Work}

\paragraph{KG}

Knowledge Graphs (KGs) are fundamental resources for knowledge-intensive NLP tasks such as dialog \citep{moon-etal-2019-opendialkg}, question answering \citep{KG-COVID-19}, named entity recognition \citep{Liu_Zhou_Zhao_Wang_Ju_Deng_Wang_2020}, open-domain questions \citep{hu-etal-2022-empowering}, and recommendation systems \citep{gao2020deep}. Recently, the target of KGs has expanded to vision and language (V\&L) fields \citep{9961954}. Based on the expansion, KGs are expected to support knowledge-intensive V\&L tasks like knowledge-intensive visual question answering \citep{yue2023mmmu}, image generation \citep{kamigaito-etal-2023-table}, explanation generation \citep{saito2024evaluating,hayashi2024artwork}, etc. In contrast to the increase in KGs' importance, the sparsity problem, which is an essential issue of KGs, still remains. As a solution, Knowledge Graph Completion (KGC) has a great role to fill in uncovered links in KGs.

\paragraph{Traditional KGC}

As introduced in \S\ref{subsec:kgc_methods}, the traditional KGC methods, represented as RESCAL \citep{rescal}, TransE \citep{transE}, DistMult \citep{distmult}, ComplEx \citep{complex}, RotatE \citep{rotate}, HAKE \citep{hake}, and HousE \citep{house} only focus on the structure of KGs, without considering the extensive textual information of KGs and pre-trained information. Thus, these models need to complete KGs only by their inference abilities. Instead of the extensive information, the modeling and training methods for the traditional KGC are well studied empirically \citep{ruffinelli2020you,ali2021pykeen} and theoretically \citep{kamigaito-hayashi-2021-unified,pmlr-v162-kamigaito22a,kamigaito2022subsampling,feng-etal-2023-model,feng2024unified} due to their simplicity. This characteristic supports the robustness and reliability of the traditional KGC.

\paragraph{PLM-based KGC}

As introduced in \S\ref{subsec:kgc_methods}, PLM-based KGC methods encode text~\cite{kg-bert, pkgc, shen-etal-2022-lass, knnkge, mem-kgc, openworld-kgc, star, wang-etal-2021-kepler, wang-etal-2022-simkgc} or generate facts~\cite{genkgc, kgt5, chen-etal-2022-s2skgc, xie-etal-2023-lambdakg, zhu2023autokg} based on pre-trained information to enhance KGC performance. There are two major categories, discrimination-based methods that utilize PLMs encoders and generation-based methods that utilize PLMs decoders~\cite{pan2023unifying}. However, it is uncertain whether the performance improvement is actually caused by the enhanced ability of inference through pre-training or data leakage from pre-trained data. We aim to reveal that in our work.

\paragraph{Data Leakage in PLMs}

Some existing datasets for the downstream tasks are often directly mixed into the pre-training data~\cite{magar-schwartz-2022-data, ml-leakage, sainz-etal-2023-nlp}, and general PLMs are not able to answer questions correctly in downstream tasks that require domain-specific knowledge excluded from the pre-trained data~\cite{wang2023on-robust-domain-specific, jullien-etal-2023-semeval, nair-modani-2023-exploiting}.

\paragraph{Inference Ability of PLMs}

Several studies~\cite{zhou-etal-2021-rica, wang2021adversarial, zhu2023autokg, zheng2023-vicuna-judging, yu2024kola, laban2023llms, qin-etal-2023-chatgpt} evaluate the inference abilities of PLMs, but they ignored the impact of the PLMs' memorization abilities in inference. Therefore, the inference abilities of PLMs remain unclear. While the memorization abilities of PLMs are beneficial~\cite{petroni-etal-2019-language, roberts-etal-2020-much, heinzerling-inui-2021-language, wei2022emergent, carlini2023quantifying-memorize}, they can introduce bias~\cite{bias1, kaneko-etal-2022-debiasing, kaneko-etal-2022-gender, meade-etal-2022-empirical, deshpande-etal-2023-toxicity, feng-etal-2023-pretraining, ladhak-etal-2023-pre} or cause errors due by the contamination in the pre-training data as hullucinations~\cite{dziri-etal-2022-origin, dziri-etal-2022-faithdial, mckenna-etal-2023-sources, hallusination-survey}. This suggests the memorization and inference abilities of PLMs are strongly related, and the pre-trained knowledge of the PLMs influences their inference abilities.

\section{Conclusion}

In this study, we proposed a method for evaluating the inference ability of PLM-based KGC methods by separately considering the information related to unseen links in KGs. Using this method as a basis, we developed synthetic datasets that focused on the structure of KGs and changed only textual information, maintaining graph structure. Then, we compared PLM-based KGC methods using these datasets.

The comparison results show that PLMs acquire the inference abilities for KGC in pre-training, whereas in KGs, they rely more on the textual information of entities and relations. Further, we observed that the KGC performance of PLM-based KGC without pre-trained knowledge is comparable to or lower than that of TransE, the traditional KGC. This highlights the importance of using both traditional and PLM-based KGC methods. 

Please see Appendix \ref{sec:exhortation} for more detailed information on improving the current KGC evaluation based on the insights from our work.

\section{Limitations}
In this study, we investigated the inference abilities of PLM-based KGC methods empirically, without focusing on theoretical verification. Furthermore, while our focus was on KGC, we did not verify whether these findings could be applied to other downstream tasks. Therefore, our future work will aim to generalize this empirical study and perform verification across various downstream tasks.

\section{Ethical Considerations}
In this study, we have created synthetic datasets derived from existing KG datasets that have cleared ethical issues following published conferences' policies.
Therefore, our created datasets do not introduce any ethical problems.

\section*{Acknowledgements}
This work was supported by JSPS KAKENHI Grant Number JP23H03458.

\bibliography{anthology,custom}

\appendix

\section{Details of PLM-based KGC Methods}
\label{sec:plm-based-methods}

\subsection{Discrimination-based Methods}
\label{subsec:discrimination-based-methods}

The early PLM-based KGC methods such as KG-BERT~\cite{kg-bert}, utilize an encoder-only PLMs like BERT~\cite{devlin-etal-2019-bert} to encode triples.
They perform binary classification to assess the plausibility of a given triplet.
KG-BERT transforms a triple $(h, r, t)$ as follows:
\begin{equation}
x=\text{[CLS]} \mathrm{Text}_h \text {[SEP]} \mathrm{Text}_r \text {[SEP]} \mathrm{Text}_t\text{[SEP]},
\end{equation}
where $\mathrm{Text}_n$ represents textual representations of $n$.
The PLM takes $x$ as input and conducts binary classification using the $\text{[CLS]}$ token $e_{[\text{CLS}]}$ from the final hidden state.
It calculates the plausibility of the triples, which is formulated as follows:
\begin{equation}
    \mathrm{Score}(h,r,t) = \mathrm{Sigmoid}(\mathrm{MLP}(e_{[\text{CLS}]})).
\end{equation}

\citet{knnkge, mem-kgc, openworld-kgc} involve filling the missing part of a triple with a $\text{[MASK]}$ token and predicting it.
The input sequence $x$ is represented as follows:
\begin{equation}
    x=\text{[CLS]} \mathrm{Text}_h \text {[SEP]} \mathrm{Text}_r \text {[SEP]} \text {[MASK]} \text{[SEP]}. \label{eq:mask-predict}
\end{equation}
Nonetheless, simply predicting the $\text{[MASK]}$ token does not facilitate direct entity prediction. Consequently, it introduces special tokens into the vocabulary to represent the corresponding entities for prediction. In the case of kNN-KGE~\cite{knnkge}, an initial learning process is undertaken when introducing these special tokens to establish the relationship between the special tokens and the entities.

The prompt shown in Equation~(\ref{eq:prompt-knnkge}) is used to mask the special tokens that represent each entity $e_i$. 
With all other parameters fixed, the masked entity $e_i$ is predicted using cross-entropy loss. 
This approach optimizes the embeddings of these entities, which are initially set to random values.
\begin{equation}
    \label{eq:prompt-knnkge}
    x_i = \text{[CLS] the description of [MASK] is } d_i \text{ [SEP]}, 
\end{equation}

Afterwards, a sentence similar to Eq.~(\ref{eq:mask-predict}) is fed into the model, which then fine-tunes the model to predict the masked entity, as formulated:
\begin{equation}
    P(t \mid h, r)=P([\text { MASK }]=\mathrm{t} \mid x, \Theta),
\end{equation}
where $\Theta$ denotes the parameters of the model.

Finally, SimKGC~\cite{wang-etal-2022-simkgc}, the state-of-the-art method employs two encoders. SimKGC splits the triple $(h, r, t)$ into a question $(h, r)$ and its answer $t$ and uses their respective PLMs to encode them into vector space, which can be expressed as:
\begin{align}
    x_{(h,r)} &= \text{[CLS] }\mathrm{Text}_h \text { [SEP] } \mathrm{Text}_r \text { [SEP] },  \label{eq:xhr}\\ 
    x_t &= \text{[CLS] } \mathrm{Text}_t\text{ [SEP] }.
\end{align}
Then, the [CLS] tokens from the final hidden state are extracted, with the embedding of $x_{(h,r)}$ represented as $e_{(h,r)}$ and the embedding of $x_t$ represented as $e_t$.
The final plausibility of the triples is scored as follows:
\begin{equation}
    \mathrm{Score} \left( (h,r), t \right) = \cos \left( e_{(h,r)}, e_t \right).
\end{equation}

Essentially, the introduced model originally employs the BERT-base model, but it can use variants of BERT such as RoBERTa~\cite{roberta}.

\subsection{Generation-based Methods}
\label{subsec:generation-based-methods}

Recently, novel KGC-based methods have been introduced that utilize Encoder-Decoder models, e.g., GenKGC~\cite{genkgc}, KGT5~\cite{kgt5}, or Decoder-only Large Language Models (LLMs), e.g., LambdaKG~\cite{xie-etal-2023-lambdakg}, AutoKG~\cite{zhu2023autokg},  to directly generate the tail entity $t$. Unlike traditional KGC methods and discrimination-based methods, which can only complete the KGs using a predefined set of entity candidates, these generation-based methods have the potential to predict unknown entities not included in the candidate list. This capability unlocks the ability to predict any and all entities in the KGs.

When predicting the missing triple $(h, t, ?)$, the model converts $x_{(h,r)}$ into a prompt specific to the models, then it into the encoder and generates $x_t$.

While there is potential to predict any and all entities, in practice, certain restrictions are put in place to focus the prediction towards entities within the KGs. For example, GenKGC introduces an entity-aware hierarchical decoder to place constraints on $x_t$.
Furthermore, KGT5 utilizes generation-based PLMs, pre-trained with text descriptions specifically for KG representation.
Notably, this is done from scratch with random initialization, rather than leveraging pre-trained models, indicating the effectiveness of a tailored approach for each dataset.\footnote{The authors mention that using pre-trained weights can improve accuracy in some cases (\url{https://github.com/intfloat/SimKGC/issues/1}). They also discuss the challenge of training models on small datasets (\url{https://github.com/apoorvumang/kgt5/issues/4}).}
Regarding the foundational models,
GenKGC employs BART-base~\cite{lewis-etal-2020-bart}, while KGT5 utilizes T5-small~\cite{t5}.

Finally, some experimental KGC methods use decoder-only LLMs. These methods employ well-designed prompts to induce in-context learning.
LambdaKG employs the information retrieval algorithm (BM25)~\cite{Buettcher2010-bm25} to construct prompts. It selects the top 100 most relevant entities from the dataset as potential answer candidates. Similarly, it retrieves the top 5 relevant triples as examples for few-shot learning. This information is aggregated into a single prompt, which is then used by LLMs to select and generate an answer.
AutoKG addresses the KGC task in a 0-shot or 1-shot setting without employing an information retrieval algorithm. It treats the missing entity as a $\text{[MASK]}$ token in the prompt and generates the corresponding value for the $\text{[MASK]}$ token using LLMs.

\begin{figure}[t]
\centering
\includegraphics[width=0.48\textwidth]{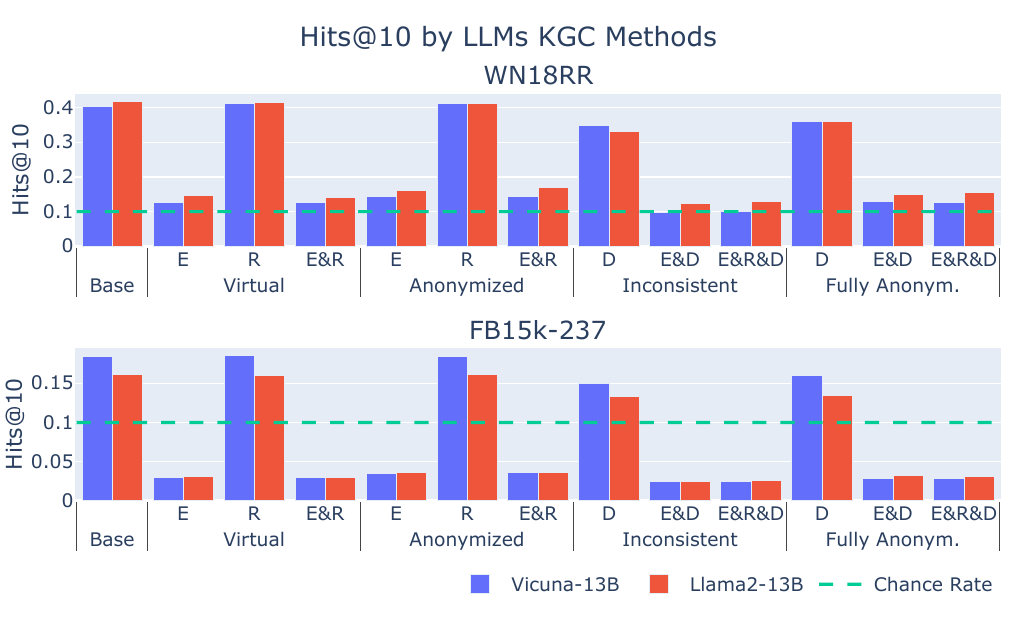}
\caption{The results of Hits@10 using vicuna-13B and Llama2-13B in the LLMs KGC methods \cite{xie-etal-2023-lambdakg}.  The LLMs select 1 entity from selected 100 candidate entities by BM25. It generates 10 sentences, and it is checked whether the correct entity is included in these. The chance rate is 0.1 because it generated total 10 entities from 100 candidates.}
\label{fig:LLMs-results}
\end{figure}

\section{Inference capabilities under a zero-shot setting with LLMs}
\label{sec:inference-capability}

We evaluate the inference capabilities in a zero-shot setting by LLMs.
We evaluate WN18RR and FB15k-237 using the LambdaKG method~\cite{xie-etal-2023-lambdakg} described in Appendix \ref{subsec:generation-based-methods}.\footnote{Original LambdaKG uses GPT-3~\cite{gpt3}, but we employ Vicuna-13B and Llama2-13B for reproducibility. These models have shown competitiveness to GPT-3 on the MT-bench Reasoning benchmark~\cite{zheng2023-vicuna-judging}. Furthermore, while the original setting calculates only Hits@1, this study calculates Hits@10 by considering the top 10 output probabilities.}
Figure \ref{fig:LLMs-results} shows the results using Vicuna-13B~\cite{zheng2023-vicuna-judging} and Llama2-13B~\cite{touvron2023llama}. The base dataset yields high hits@10 scores, but when entities are changed, the impact is high, and it is small when only descriptions are changed. However, LLMs don't know how the entity was changed, so the chance rate serves as an upper limit. Therefore, it is clear that inference by LLMs is based on pre-trained knowledge.

\section{Exhortation to KGC}
\label{sec:exhortation}

\paragraph{Datasets}

As discussed in Section \ref{sec:which-factors}, the information for relations has very little impact. Some entities are assigned only one relation, as shown in Table \ref{tab:dataset_stats}. Thus, if only the entity is known, it may be possible to infer the unknown entities without relation information. Traditional KGC methods without PLMs can learn the graph structure from scratch. In contrast, PLMs' knowledge can help with completion without relation information, as discussed in Section \ref{sec:inference-capability}. The current dataset focuses on entities, but it cannot accurately measure the effect of relations. Therefore, a dataset that specifically focuses on relations is needed.

Next, according to Table \ref{tab:entity_in_descriptions}, it has become clear that the missing entity information is included in the descriptions of queries. Therefore, if we use descriptions in the KGC task, it can be considered a cheat setting, as it utilizes the information extraction capability from the text data in PLMs. The descriptions are indeed useful for disambiguation in entities, but they also provide too much information for inference, thus demonstrating information extraction capabilities. In the future, to measure the pure inference capabilities for unknown entities, descriptions should not be used in the KGC task for fair comparison.

\paragraph{Models}
As discussed in Section \ref{sec:The-Effects-of-PLMs-Knowledge}, PLMs' knowledge helps inferences for unknown entities. 
Therefore, when we evaluate filling in truly unknown links in KGs by KGC in the future, 
we should avoid using pre-trained weights. This suggests that PLM-based KGC methods with pre-trained weights create a cheat setting because they utilize external knowledge not included in datasets, which does not measure the pure inference capabilities for unknown entities in KGC tasks. It is essential to evaluate the model's performance based on the target KGC dataset only for a fair comparison.




\end{document}